\documentclass{article}

\usepackage{microtype}
\usepackage{graphicx}
\usepackage{subcaption}
\usepackage{booktabs}
\usepackage{multirow}
\usepackage{makecell}
\usepackage{xcolor}
\usepackage{colortbl}
\usepackage{enumitem}

\usepackage{hyperref}
\usepackage{longtable}
\usepackage{array}
\usepackage{tcolorbox}
\usepackage{listings}
\newcommand{\RComment}[1]{\hfill {\color{gray}\footnotesize // #1}}

\usepackage[preprint]{icml2026}

\usepackage{amsmath}
\usepackage{amssymb}
\usepackage{mathtools}
\usepackage{amsthm}

\usepackage[capitalize,noabbrev]{cleveref}

\theoremstyle{plain}

\theoremstyle{definition}

\theoremstyle{remark}

\usepackage[textsize=tiny]{todonotes}

\newcommand{\zy}[1]{#1}

\begin{document}

\twocolumn[
\icmltitle{CARL: Criticality-Aware Agentic Reinforcement Learning}

\icmlsetsymbol{equal}{*}

\begin{icmlauthorlist}
\icmlauthor{Leyang Shen}{nus}
\icmlauthor{Yang Zhang}{nus}
\icmlauthor{Chun Kai Ling}{nus}
\icmlauthor{Xiaoyan Zhao}{nus}
\icmlauthor{Tat-Seng Chua}{nus}
\end{icmlauthorlist}

\icmlaffiliation{nus}{National University of Singapore, Singapore}

\icmlcorrespondingauthor{Yang Zhang}{zhangy@nus.edu.sg}

\icmlkeywords{Machine Learning, ICML}

\vskip 0.3in
]

\printAffiliationsAndNotice{}

\begin{abstract}

Agents capable of accomplishing complex tasks through multiple interactions with the environment have emerged as a popular research direction. However, in such multi-step settings, the conventional group-level policy optimization algorithm becomes suboptimal because of its underlying assumption that each step holds equal contribution, which deviates significantly from reality. Our analysis reveals that only the action choices on a small fraction of states are critical in determining the final outcome. Building on this insight, we propose CARL, a criticality-aware reinforcement learning algorithm tailored for long-horizon agentic reasoning. CARL leverages entropy as a heuristic proxy for state criticality and achieves focused training by assigning rewards to actions taken from high-criticality states while excluding actions taken from low-criticality states from model updates, avoiding noisy credit assignment and redundant computation. Extensive experiments demonstrate that CARL achieves both stronger performance and higher efficiency across diverse evaluation settings. The source code will be publicly available.

\end{abstract}

\section{Introduction}

Large Language Model (LLM)-based agents have witnessed significant development~\cite{wei2026agentic}. With powerful reasoning and tool usage capabilities, 
agents can autonomously plan and interact with their environment in a goal-oriented manner to accomplish complex tasks~\cite{xi2025rise, wang2024survey, kong2025token}.
Among them, multi-turn search agents~\cite{jin2025search-r1, gao2025asearcher} represent one of the most advanced directions,
which can engage in multiple rounds of interaction with search engines and browsers to gather useful information for addressing knowledge-intensive questions~\cite{yang2018hotpotqa, wikimultihop}. This interactive mechanism marks a shift from single-step inference to multi-step, feedback-driven execution, positioning it as a high-value research area.

Reinforcement learning (RL) plays a crucial role in enhancing \textit{multi-turn search agents}, as it enables them to self-improve without relying on human supervision. However, most of the existing attempts on search agent RL~\cite{jin2025search-r1, gao2025asearcher, song2025r1-searcher} leverage group-level policy optimization (GRPO)~\cite{shao2024deepseekmath} algorithm without examining its suitability.
GRPO assumes that each part in a trajectory contributes equally to the outcome~\cite{tan2025gtpo} and repeatedly rolls out full trajectories from scratch, suffering from noisy credit assignment and redundant computation. These issues become particularly pronounced in long-horizon agentic tasks~\cite{gao2025asearcher}, where trajectories span multiple steps yet only a small fraction of actions are decisive.

\begin{figure}
    \centering
    \includegraphics[width=1.0\linewidth]{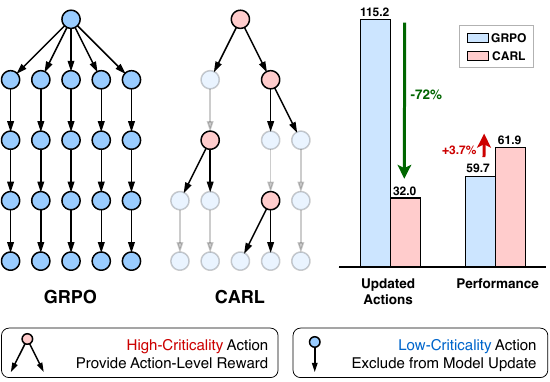}
    \caption{GRPO repeatedly rolls out full trajectories from scratch, suffering from noisy credit assignment and redundant computation. CARL addresses these issues by focusing learning exclusively on high-criticality actions, achieving higher performance while updating the policy on 72\% fewer actions.}
    \label{fig:Intro-small}
\end{figure}

We perform a preliminary study that supports this observation, showing that actions at different steps play different roles and have varying impacts on the final outcome. Specifically, when resampling the actions at individual states, over half of the states induce near-zero changes in the final reward, whereas only a small subset could cause sharp reward changes.
These findings suggest that states differ in their criticality.

In light of this, we argue that RL for agents should focus on actions from high-criticality states, and accordingly propose the \textit{\textbf{C}riticality-\textbf{A}ware \textbf{R}einforcement \textbf{L}earning (CARL)} algorithm that enjoys both high performance and training efficiency.
CARL leverages action entropy as a proxy for criticality, concentrating rollouts and optimization efforts on high-criticality states rather than distributing resources uniformly.
As illustrated in Fig.~\ref{fig:Intro-small}, CARL initiates rollouts only from high-criticality states and assigns action-level rewards based on expected reward gains, while excluding actions taken from low-criticality states from gradient updates.
This targeted focus effectively mitigates noisy credit assignment and redundant computation inherent in GRPO.

We follow the setting of ASearcher~\cite{gao2025asearcher} and evaluate CARL on knowledge-intensive question-answering (QA) tasks. As shown in Fig.~\ref{fig:Intro-small}, CARL achieves superior accuracy while requiring policy updates on 72\% fewer actions than GRPO. These results demonstrate that CARL is an effective RL algorithm for agentic reasoning that delivers both performance and training efficiency advantages. 

Our main contributions are summarized as follows: 
\begin{itemize}[leftmargin=*, itemsep=0.4em, parsep=0pt, topsep=-3pt]
    \item We provide the first comprehensive analysis of the multi-turn search agent pipelines, revealing that only the action choices at a small subset of states have high impact on the final outcome, which motivates focusing optimization on actions taken from these critical states.
    \item We design CARL, a reinforcement learning framework tailored to agentic reasoning. This framework performs focused training on actions taken from high-criticality states, yielding both high performance and efficiency.
    \item We conduct comprehensive experiments on multi-turn search agents across different model sizes and both reasoning and non-reasoning models, demonstrating consistent improvements on multiple knowledge QA benchmarks.
\end{itemize}

\section{Related Works}

In this section, we discuss recent research advances in two related fields: search agents and improving RL for LLM through credit assignment.

\subsection{Search Agent}

With the rapid advancement of LLMs’ core capabilities~\cite{qwen3, guo2025deepseek}, search agents~\cite{yao2022react} \zy{have emerged by} equipping LLMs with \zy{internet access} and custom workflows. These agents can proactively search and browse the web to gather information before responding. As task complexity increases, static prompt engineering~\cite{li2025search-o1, xinjie2025reagent} and dataset construction method~\cite{yu2024auto-rag} quickly reach their limits in improving performance.
\zy{In contrast, reinforcement learning offers greater potential by enabling agents to self-improve through interaction with the environment.}

Inspired by the success of GRPO~\cite{shao2024deepseekmath} on math reasoning~\cite{hendrycks2021math, he2024olympiadbench}, recent studies have attempted to extend it to search agents~\cite{song2025r1-searcher, jin2025search-r1, zheng2025deepresearcher, gao2025asearcher}, with a primary focus on data synthesis and framework construction.
\zy{ARPO~\cite{dong2025arpo} takes an initial step toward RL algorithmic refinement by shifting the rollout granularity from trajectory to step level,}
based on the observation that token entropy \zy{often} rises after tool calls.
However, these methods still treat all steps uniformly without examining their individual contributions, leading to noisy credit assignment and redundant computation.
In this work, we discover that agent outcomes are determined by the action choices at a small subset of high-criticality states, and accordingly structure RL to focus on actions taken from these states.

\subsection{Credit Assignment in Reinforcement Learning}

Many recent works~\cite{wang2023math-shepherd,luo2024improve,setlur2024rewarding,wang2025spa} enhance RL for LLM through finer-grained credit assignment, as step-wise rewards can accelerate convergence and improve learning outcomes~\cite{lightman2023lets_verify}. Among these, reward-model-free methods~\cite{kazemnejad2024vineppo,fei2025SPRO} stand out for their stability and efficiency. 

A natural structure for providing fine-grained, step-level rewards is a tree\footnote{Unlike Monte Carlo tree search (MCTS)~\cite{browne2012mcts}, which focuses on iteratively searching for complete successful paths, tree-based RL methods instead leverage tree structures for rollouts to estimate step-level relative advantages, without emphasis on successful paths.}, enabling controlled comparisons among intermediate steps~\cite{tran2025exploiting}. 
Prior studies have explored this structure on math reasoning tasks. For example, TreeRPO~\cite{yang2025treerpo} samples a full $n$-ary tree and groups sibling nodes for GRPO updates. TreeRL~\cite{hou2025treerl} further removes the group constraint and directly estimates each step's advantage by evaluating its impact on expected rewards. In agentic RL~\cite{gao2025asearcher}, reward sparsity becomes more critical due to long-horizon multi-step execution, as demonstrated by ReasonRAG~\cite{zhang2025process}, \zy{rendering step-level reward modeling increasingly necessary}.
\zy{In this work, we address this and propose assigning rewards exclusively to actions taken from high-criticality states that truly determine the outcome.}

\begin{table}[t]
    \small
    \centering
    \caption{\textbf{Example of a Search Agent Execution Pipeline.} The agent interacts with the environment multiple times to gather useful information, with each action playing different roles.}
    \label{tab:method-example}
    \begin{tabular}{lp{0.76\linewidth}}
        \toprule
        \multicolumn{2}{c}{\textbf{Where did Ruth Scurr's husband graduate from?}} \\
         \midrule
        Search & Ruth Scurr husband university \\
        \midrule
        Read & ... The most relevant link from the results is the Wikipedia page for Ruth Scurr (Doc 4), "she was married to the political theorist John Dunn", but provides no information about his education ... \\
        \midrule
        Search & John Dunn education \\
        \midrule
        Read & ... I've confirmed that Ruth Scurr's husband is John Dunn (the political theorist), born in 1940 ... need to access this page for detailed information.  \\
        \midrule
        Access & ... title=John\_Dunn\_(political\_theorist) \\
        \midrule
        Read & ... the page states: "... He read history at King's College, Cambridge ..." This confirms that Ruth Scurr's husband, John Dunn (the political theorist), graduated from King's College, Cambridge with a degree in history. \\
        \midrule
        Answer & King's College, Cambridge \\
        \bottomrule
    \end{tabular}
\end{table}

\section{Preliminaries}
Multi-turn search agents~\cite{gao2025asearcher} are equipped with a search engine and a web browser. They are designed to answer knowledge-intensive questions through interacting with these tools several times for information gathering.
We formulate this execution process as a Markov Decision Process (MDP), defined by the tuple $\langle \mathcal{S}, \mathcal{A}, P, R \rangle$, where $\mathcal{S}$ represents the state space, and $\mathcal{A}$ denotes the action space. $P(s_{t+1} \mid s_t,a)$ is the transition probability from state $s_t$ to $s_{t+1}$ after taking action $a$, which represents the tool calling process in practice.
$R(s)$ is a rule-based outcome reward function, defined to be non-zero only at terminal states $s_T$.

The agent is represented by a policy $\pi_\theta(a \mid s)$, which is defined by a probability distribution over actions given a state.
In practice, the LLM-based agent chooses actions in language space by generating a sequence of tokens $a = (a^{1}, \ldots, a^{|a|})$ autoregressively from a finite vocabulary $\mathcal{V}$, i.e., $\pi_\theta(a \mid s) = \prod_{j=1}^{|a|} \pi_\theta(a^{j} \mid s, a^{<j})$.
For each question, the agent will continuously generate $a_t$ for $s_t$ until reaching a termination state $s_T$ with an ``answer'' or reaching the maximum step limit $T_{\max}$. The whole execution process can be recorded as a trajectory $\tau = \{(s_t, a_t)\}_{t=1}^{T}$.
The objective of RL is to maximize the outcome reward, which can be expressed as
\begin{equation}
J(\pi_\theta) = \mathbb{E}_{\tau\sim(\pi_\theta,P)} [R(\tau)].
\end{equation}
GRPO~\cite{shao2024deepseekmath} follows the widely-adopted online RL algorithm, PPO~\cite{schulman2017ppo}, to setup loss function $\mathcal{J}(\theta)$:
{\small
\begin{equation}
\label{eq:ppo}
\mathcal{J}(\theta)
=
\frac{1}{|\mathcal{D}|}
\sum_{(s_i,a_i,A_i)\in\mathcal{D}}
\min\Big[
    r_i(\theta)A_i,
    \mathrm{clip}\big(r_i(\theta),1-\epsilon,1+\epsilon\big)A_i
\Big],
\end{equation}
\begin{equation}
r_i(\theta)
=
\frac{\pi_\theta(a_i\mid s_i)}{\pi_{\theta_{\text{old}}}(a_i\mid s_i)},
\end{equation}}
where $\mathcal{D}$ denotes the dataset, $\pi_\theta$ is the policy parameterized by $\theta$. $A_i$ is the advantage of token $i$, which indicates the direction and scale the model should update on it. The ratio $r_i(\theta)$ measures the policy shift between the updated policy $\pi_\theta$ and the old policy $\pi_{\theta_{\text{old}}}$, and is clipped by $\epsilon$ for stability.

GRPO measures the advantage of each trajectory $\tau$ by comparing it with the group average:
\begin{equation}
    A_\tau = \frac{R(\tau) - \text{mean}(R(\tau))}{\text{std}(R(\tau))}
\end{equation}
where $R(\tau)$ are rewards of trajectories based on their terminal states, while $\text{mean}(R(\tau))$ and $\text{std}(R(\tau))$ are computed
over all trajectories of the same question. This trajectory-level advantage is then assigned to all tokens within the trajectory under the assumption that every token contributes equally.

\section{Agentic Reasoning Pipeline Analysis}\label{sec:analysis}
GRPO's assumption of equal token contribution becomes problematic in agentic reasoning, where trajectories consist of clearly separated steps that play different roles and with varying criticality.
We analyze the reasoning pipeline through a preliminary experiment to characterize this variation in state criticality, informing our algorithm design.

\paragraph{Case Study.}
In Table~\ref{tab:method-example}, we present an example of how a multi-turn search agent solves a problem. The agent's actions can be categorized into \emph{search}, \emph{access}, \emph{read}, and \emph{answer}. In the \emph{search} action, the agent generates search keywords and gets results from a search engine. In the \emph{access} action, it accesses a specific website by URL. After each of them, the agent performs \emph{read} actions to extract and summarize useful information from the often lengthy search results or webpage content. Once sufficient evidence is collected, the agent executes the \emph{answer} action, terminating the pipeline with a final response. Intuitively, these actions play different roles and contribute differently to the outcome.

\paragraph{State Criticality Distribution.}
To validate this intuition, we conduct an experiment to observe the distribution of state criticalities.
We define the \emph{criticality} of a state, denoted $C_{\pi_\theta}(s_t)$, as the degree to which the action choice taken at this state influences the final outcome. Formally, we quantify criticality as the standard deviation in outcome reward when stochasticity is isolated to the action sampled at that state:
\begin{equation}\label{eq:criticality}
      C_{\pi_\theta}(s_t) = \mathrm{std}_{a' \sim \pi_\theta(\cdot | s_t)}\Big[ R(\tau_{s_t,a'}) \Big]
\end{equation}
where $\tau_{s_t,a'}$ denotes the trajectory taking action $a'$ at state $s_t$ followed by greedy decoding thereafter (\textit{i.e.,} temperature=0).

\begin{figure}[t]
    \centering
    \includegraphics[width=1.0\linewidth]{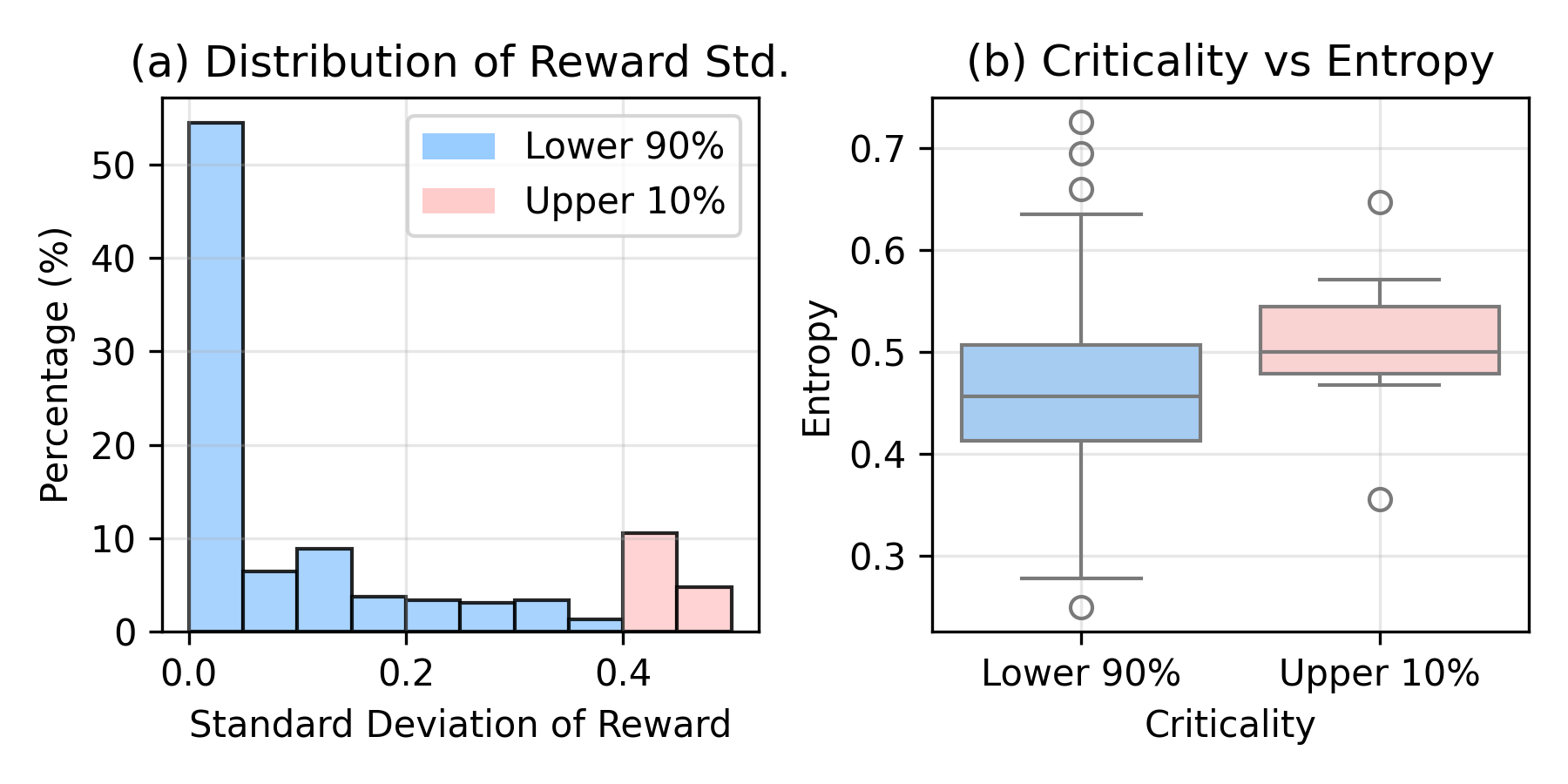}
    \caption{\textbf{Quantitative Analysis of Execution Pipeline.} (a) Most actions yield low reward variance when resampled, while only a small subset exhibits notably high variance. (b) The states corresponding to high-criticality actions show higher entropy than those associated with low-criticality actions.}
    \label{fig:importance_analysis}
\end{figure}

We test the search agent on 70 tasks randomly sampled from the 7 knowledge-intensive question-answering datasets used in our evaluation (Section~\ref{sec:setup}), resulting in 294 states. For each state, we estimate $C_{\pi_\theta}(s_t)$. As shown in Fig.~\ref{fig:importance_analysis}(a), we can observe substantial disparity in state criticality. Most states exhibit extremely low variance in reward: over 50\% of states have almost no impact on the outcome when their actions are resampled, indicating that the majority of states are of low criticality.
In contrast, around 10\% of states exhibit a reward standard deviation greater than 0.4, representing a minority of high-criticality states.

This suggests that uniform treatment of all actions, as done by GRPO, is suboptimal. Instead, we should take the criticality of states into account: assigning rewards more precisely to actions taken from high-criticality states and concentrating computational resources on them.

\section{Criticality-Aware RL}~\label{sec:algo}
\begin{figure*}
    \centering
    \includegraphics[width=1.0\linewidth]{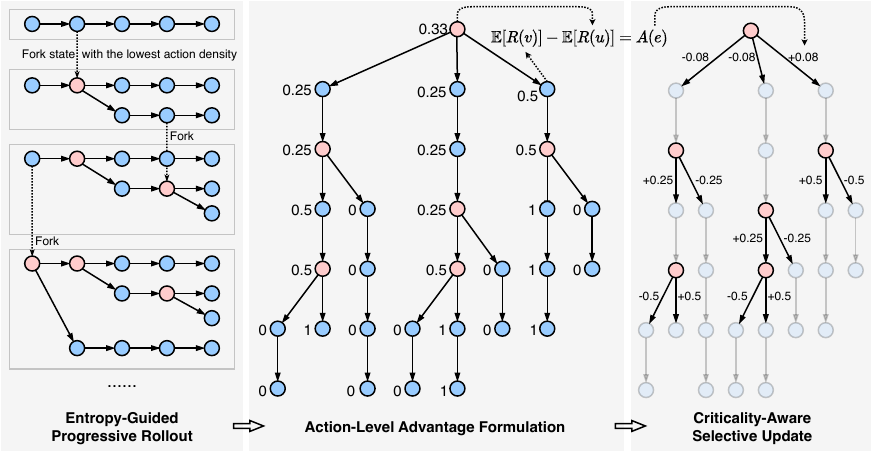}
    \caption{\textbf{CARL Algorithm.} In the rollout phase, CARL progressively forks the state with the lowest action density. Then, it assigns action-level credits to critical actions through an expected-reward-gain formulation: the expected reward of each state is estimated by averaging its successor states, and the advantage of an action is computed as the difference between the terminal and initial state. In the model update phase, low-criticality actions are excluded (grayed out) to reduce redundant computation.}
    \label{fig:main-method}
\end{figure*}

Inspired by the study, we propose the \textit{Criticality-Aware RL} (CARL) algorithm. As illustrated in Fig.~\ref{fig:main-method}, CARL prioritizes exploration at high-criticality states with entropy-guided progressive rollout, and provides action-level update signals using action-level advantage formulation. Then, it performs selective updates on actions from high-criticality states, excluding those from low-criticality ones.

\paragraph{Critical States Identification.}
Before we can focus on critical states,
we need a practical method to identify them, as the sampling-based estimation in Eq.~\ref{eq:criticality} incurs a high computational cost. To achieve this, we analyze the characteristics of high-criticality states to find an practical metric.

High-criticality states are intuitively non-trivial decision points where multiple candidates appear plausible, making it difficult to identify the optimal choice. This difficulty can be reflected in the model's action distribution: when the model cannot confidently make decisions, it assigns similar probabilities to multiple candidates, resulting in a more uniform distribution. We adopt entropy to measure this quantitatively, termed \emph{action entropy}, which captures the degree to which probability mass is dispersed across possible actions. A higher action entropy at state $s_t$ indicates greater uncertainty in selecting among competing actions $a_t$, suggesting that $s_t$ is of higher criticality.

Unlike token-level entropy, which can be calculated precisely by enumerating over a finite vocabulary, action entropy involves an infinite space of possible actions and thus requires estimation. 
We estimate the action entropy at state $s_t$ via Monte Carlo sampling~\citep{robert1999monte}:
\begin{align}\label{eq:uncertainty}
\hspace{-0.5em}
H_{\pi_\theta}(s_t)
&= \mathbb{E}_{Y\sim \pi_\theta(\cdot\mid s_t)}
   \bigl[-\log \pi_\theta(Y\mid s_t)\bigr] \\
&\approx \frac{1}{N}\sum_{i=1}^N
   \bigl[-\log \pi_\theta(Y^{(i)}\mid s_t)\bigr],
    Y^{(i)}\sim \pi_\theta(\cdot\mid s_t),
\end{align}
where $Y^{(i)}$ denotes an action sampled by policy $\pi_\theta$ at state $s_t$, and $N$ denotes the total sample size.
Each sampled action $Y^{(i)}$ is a sequence of tokens, $Y^{(i)} = [y^{(i)}_1,\dots, y^{(i)}_{|Y^{(i)}|}]$, with $|Y^{(i)}|$ denoting its length. We follow \citet{wu2016google} and adopt a length-normalized sequence log-probability:
\begin{equation}
    \log \pi_\theta(Y^{(i)}\mid s_t)=\frac{1}{|Y^{(i)}|}\sum_{j=1}^{|Y^{(i)}|} \log \pi_\theta(y^{(i)}_j\mid y^{(i)}_{<j},s_t).
\end{equation}
This entropy-based criticality measure aligns with intuition: we typically feel uncertain when facing decisions that significantly affect outcomes, hesitating among multiple plausible options rather than acting reflexively.
Our empirical results confirm this: as shown in Fig.~\ref{fig:importance_analysis}(b), high-criticality states exhibit clearly higher action entropy than low-criticality states. This difference is statistically significant (Brunner--Munzel test, $p=0.002$) with a medium-to-large effect size (Cliff's $\delta=0.42$).
This finding validates action entropy as an effective proxy for criticality, eliminating the need for costly outcome-based sampling.

\paragraph{Action-Level Advantage Formulation.}
To provide precise update signals for the actions taken at the identified high-criticality states, CARL assigns rewards to them with an action-level advantage formulation.
We move beyond the group-based RL paradigm of GRPO and reformulate the credit assignment question from ``How good is this trajectory compared to group average?'' to ``How much reward improvement does this action provide?''.

Specifically, we organize sampled trajectories into a tree structure, where nodes represent states and directed edges represent actions, as shown in Fig.~\ref{fig:main-method}. We first compute the expected reward of each state via a recursive Bellman-style estimator over this tree and then calculate the advantage of each action through tree differencing, which measures the expected reward gain contributed by taking that action.
Formally, the expected reward $\mathbb{E}[R(u)]$ of a state $u$ is defined as the expected outcome reward obtained by following the policy $\pi_\theta$ starting from the state $u$:
\begin{equation}
    \mathbb{E}[R(u)]
    = \mathbb{E}_{\pi_\theta}\Big[ R(\tau) \,\big|\, s_0 = u \Big].
\end{equation}
This can be recursively computed by averaging its child nodes $v_i \sim \pi_\theta(\cdot \mid u)$:
\begin{align}
\hspace{-0.2em}
\mathbb{E}[R(u)]
= \sum_{v} p(v \mid u)\,\mathbb{E}[R(v)]
\approx
\frac{1}{N}\sum_{i=1}^{N} \mathbb{E}[R(v_i)].
\end{align}
For each action, represented by edge $e=(u,v)$ connecting parent node $u$ and child node $v$, the advantage is defined as
\begin{align}\label{eq:advantage}
A(e) = \mathbb{E}[R(v)] - \mathbb{E}[R(u)]
\end{align}
Intuitively, $A(e)$ measures how much taking action $e$ improves the expected outcome compared to state $u$. This action-level advantage formulation provides an unbiased estimation of action-level reward regardless of tree structure. This unbiasedness does not require a full $n$-ary tree, requiring only that child nodes are sampled independently from $\pi_\theta(\cdot\mid u)$. This property distinguishes CARL from prior tree-based methods such as ARPO~\cite{dong2025arpo} and TreeRL~\cite{hou2025treerl}, as analyzed in Appendix~\ref{sec:unbiased}.

\paragraph{Entropy-Guided Progressive Rollout.}
We leverage the flexibility brought by unbiasedness to dynamically allocate rollout budget in proportion to each state's action entropy ( Eq.~\eqref{eq:uncertainty}),
thereby maximizing the criticality of explored states. Specifically, we define the \emph{action density} to measure the extent to which each state has been explored relative to its entropy:
\begin{equation}
d(s_t) =\frac{n(s_t)}{H_{\pi_\theta}(s_t)},
\end{equation}
where $n(s_t)$ denotes the number of children already sampled from $s_t$. 
At each expansion step, CARL greedily selects the node with the lowest action density to start with, as illustrated in Fig.~\ref{fig:main-method}.

In practice, we set an initial sample size $N_0$, a hyperparameter, to control how many trajectories should be generated from scratch before the forking algorithm. This design improves stability by ensuring a basic set of candidates to start with. We present the progressive rollout algorithm in Appendix~\ref{app:rollout}.

\paragraph{Criticality-Aware Selective Update.}

As a result of entropy-guided rollout, $\mathcal{D}_\text{roll}$ comprises state-action pairs of two kinds: at states identified as high-criticality, the algorithm branches and samples multiple actions, yielding \emph{siblings}---alternative actions sampled from the same parent state; at states identified as low-criticality, only a single action is sampled, so the action has no sibling.
Each training sample is a tuple of state $s_t$, action $a_t$, and advantage $A_t$. As actions from low-criticality states have little impact on the final results, we exclude their data from model update to further improve efficiency and avoid incorrect learning. The resulting updating sample set $\mathcal{D}_{\text{upd}}$ is
\begin{equation}
\mathcal{D}_{\text{upd}} = \{(s_t, a_t, A_t)\mid (s_t,a_t,A_t)\in\mathcal{D}_\text{roll}, |\text{child}(s_t)|>1\}.
\end{equation}
For each action, we follow the setting of GRPO to assign the action-level reward to all tokens within the action.
The model is then updated on $\mathcal{D}_{\text{upd}}$ using the PPO loss (Eq.~\ref{eq:ppo}).

\paragraph{Efficiency Analysis.}
CARL improves efficiency by reducing computing resource consumption for both phases of RL: trajectory collection and model update.
For trajectory collection, we reuse the prefix of existing sequences without repeatedly rolling out from scratch. Under the condition that the number of leaf nodes $N$ remains unchanged, the total number of actions required is significantly reduced. Assume each trajectory contains $T$ actions and the probability of an action being critical is uniformly distributed. The expected resource consumption as a proportion of the baseline equals
\begin{equation}
    \frac{TN_0 + \frac{T}{2}N}{NT} = \frac{N_0}{N} + \frac12.
\end{equation}
When $N_0$ and $N$ are set to $1$ and $16$ respectively in our default setting, CARL saves $44\%$ of resources. When we increase $N_0$ to $8$ in our best performance setting, CARL keeps the same rollout resource consumption as GRPO.

$|\mathcal{D}_\text{upd} |$

For model update, the sample size $|\mathcal{D}_\text{upd} |$ is further reduced by excluding actions from low-criticality states. Since forking from a state without siblings adds two new actions to the update set, while forking from a state already with siblings adds only one, the number of actions used for updating is bounded by
\begin{equation}
N+1 \le |\mathcal{D}_\text{upd} | \le 2N.
\end{equation}
In contrast, GRPO uses $TN$ action samples per question, where statistics show $\mathbb{E}[T]=5$. 
Therefore, CARL uses 60\% fewer actions for model update, substantially improving training efficiency.

\begin{table*}[t]
\centering
\caption{\textbf{Results on Knowledge-Intensive Question Answering Benchmarks.} $|\mathcal{D}_\text{roll}|$ Avg. and $|\mathcal{D}_\text{upd}|$ Avg. denote the average number of actions performed during rollout phase and samples used for model update per task, respectively. Lower values indicate reduced computational cost. The best results are in bold.}
\label{tab:local-kb}
\resizebox{\textwidth}{!}{
\begin{tabular}{l|cccccccc|cccccc|cc|cc}
\toprule
\multirow{3}{*}{Method} 
& \multicolumn{8}{c|}{Multi-Hop QA $\uparrow$} 
& \multicolumn{6}{c|}{Single-Hop QA $\uparrow$} 
& \multicolumn{2}{c|}{Avg. $\uparrow$}
& \multicolumn{2}{c}{Efficiency $\downarrow$} \\
\cmidrule(lr){2-9} \cmidrule(lr){10-15} \cmidrule(lr){16-17} \cmidrule(lr){18-19}
& \multicolumn{2}{c}{2WikiMQA} 
& \multicolumn{2}{c}{HotpotQA} 
& \multicolumn{2}{c}{Bamboogle} 
& \multicolumn{2}{c|}{Musique} 
& \multicolumn{2}{c}{NQ} 
& \multicolumn{2}{c}{TriviaQA} 
& \multicolumn{2}{c|}{PopQA} 
& \multirow{2}{*}{F1}
& \multirow{2}{*}{LasJ}
& \multirow{2}{*}{\makecell[c]{$|\mathcal{D}_\text{roll}|$\\Avg.}} 
& \multirow{2}{*}{\makecell[c]{$|\mathcal{D}_\text{upd}|$\\Avg.}}
 \\
& F1 & LasJ & F1 & LasJ & F1 & LasJ & F1 & LasJ
& F1 & LasJ & F1 & LasJ & F1 & LasJ
&  &  &  &  \\
\midrule
\multicolumn{19}{c}{\textbf{3B Non-Reasoning Models (max 32 actions)}} \\
\midrule
Zero-Shot 
& 16.1 & 22.9 & 16.4 & 26.8 & 19.0 & 23.2 & 8.4 & 8.1 
& 19.4 & 30.5 & 27.6 & 50.5 & 19.3 & 25.8 
& 18.0 & 26.8
& - & - \\
GRPO 
& 27.8 & \textbf{27.1} & 33.7 & 34.8 & 24.5 & 20.8 & \textbf{14.3} & \textbf{10.7} 
& 42.9 & 40.8 & 50.1 & 56.9 & 39.9 & 36.7 
& 33.3 & 32.5
& 32.9 & 32.9 \\
\rowcolor{blue!10}
CARL
& \textbf{29.2} & 26.7 & \textbf{34.9} & \textbf{35.8} & \textbf{25.7} & \textbf{22.4} & 13.7 & 10.1 
& \textbf{43.1} & \textbf{40.9} & \textbf{50.6} & \textbf{58.5} & \textbf{40.6} & \textbf{37.9} 
& \textbf{34.0} & \textbf{33.2}
& \textbf{32.1} & \textbf{32.0} \\
\midrule
\multicolumn{19}{c}{\textbf{7B Non-Reasoning Models (max 32 actions)}} \\
\midrule
Zero-Shot 
& 17.3 & 30.8 & 17.4 & 36.7 & 16.5 & 36.8 & 9.2 & 12.9
& 22.0 & 45.4 & 26.3 & 61.3 & 20.5 & 36.9
& 18.5 & 37.3
& - & -\\
GRPO 
& 49.2 & 51.9 & 47.0 & 48.7 & 40.7 & 44.8 & 26.5 & 23.0
& \textbf{51.9} & \textbf{52.1} & 60.2 & 70.6 & \textbf{48.8} & \textbf{47.3}
& 46.3 & 48.3
& \textbf{55.3} & 55.3\\
\rowcolor{blue!10}
CARL
& \textbf{54.5} & \textbf{57.6} & \textbf{51.9} & \textbf{54.6} & \textbf{48.6} & \textbf{49.6} & \textbf{27.1} & \textbf{23.7}
& 49.1 & 50.0 & \textbf{61.5} & \textbf{72.2} & 46.8 & 45.6
& \textbf{48.5} & \textbf{50.5}
& 82.3 & \textbf{31.2} \\
\midrule
\multicolumn{19}{c}{\textbf{4B Reasoning Models (max 10 actions)}} \\
\midrule
Zero-Shot       
& 41.5 & 54.3 & 31.9 & 47.4 & 28.5 & 46.4 & 4.1 & 19.6
& 39.6 & 46.8 & 55.0 & 69.6 & 45.7 & 47.3
& 35.2 & 47.3
& - & -\\
GRPO            
& \textbf{70.9} & \textbf{73.4} & 60.7 & 64.4 & 55.1 & 58.4 & 33.4 & 32.5
& 54.9 & 54.5 & 68.6 & 79.5 & 55.2 & \textbf{53.5}
& 57.0 & 59.5
& 80.9 & 80.9 \\
TreeRPO
& 70.1 & 72.6 & 59.6 & 64.1 & 55.7 & 57.6 & 32.4 & 31.2
& 55.4 & 54.1 & 68.7 & 77.6 & 53.4 & 50.5
& 56.5 & 58.2
& \textbf{36.7} & 36.7 \\
TreeRL
& 69.5 & 71.5 & \textbf{61.1} & 64.3 & 57.7 & 58.4 & 32.8 & 30.5
& 56.3 & 56.7 & 69.8 & 79.6 & 55.8 & 53.3
& 57.6 & 59.2
& 52.1 & 52.1\\
ARPO 
& 68.7 & 70.4 & 60.9 & \textbf{65.3} & 59.6 & 61.6 & 32.6 & 31.2 
& 55.7 & 56.1 & \textbf{70.8} & \textbf{81.6} & 55.3 & 52.7 
& 57.7 & 59.8 
& 57.2 & 57.2 \\
\rowcolor{blue!10}
CARL-Lite            
& 69.1 & 71.8 & 59.2 & 62.9 & \textbf{61.1} & 61.6 & \textbf{34.6} & \textbf{33.0}
& 55.5 & 56.4 & 70.1 & 80.1 & 54.0 & 51.7
& 57.6 & 59.6
& 39.8 & \textbf{30.5} \\
\rowcolor{blue!10}
CARL            
& 70.6 & 72.2 & 60.8 & 65.1 & 60.7 & \textbf{63.2} & 33.7 & 32.3
& \textbf{57.0} & \textbf{58.6} & 69.7 & 79.9 & \textbf{56.2} & \textbf{53.5}
& \textbf{58.4} & \textbf{60.5}
& 81.8 & 32.0 \\
\midrule
\multicolumn{19}{c}{\textbf{4B Reasoning Models (max 32 actions)}} \\
\midrule
GRPO & 71.1 & 73.5 & 59.3 & 63.1 & 59.5 & 63.2 & 32.0 & 31.1 & 55.5 & 55.0 & 69.2 & 78.6 & 54.9 & 53.1 & 57.4 & 59.7
& \textbf{115.2} & 115.2 \\
\rowcolor{blue!10}
CARL & \textbf{74.0} & \textbf{77.3} & \textbf{62.6} & \textbf{66.9} & \textbf{61.6} & \textbf{64.0} & \textbf{33.9} & \textbf{33.5} & \textbf{55.8} & \textbf{56.3} & \textbf{70.8} & \textbf{81.8} & \textbf{55.7} & \textbf{53.5} & \textbf{59.2} & \textbf{61.9}
& 141.3 & \textbf{32.0} \\
\bottomrule
\end{tabular}
}
\end{table*}

\section{Experiments}

\subsection{Experimental Setup}\label{sec:setup}

We follow the experimental settings of ASearcher~\cite{gao2025asearcher}, using a local retrieval server as search environment, which leverages the 2018 Wikipedia dump~\cite{karpukhin2020dense} as information source and E5~\cite{wang2022e5} as the retriever. We evaluate on seven knowledge QA benchmarks, spanning single-hop (Natural Questions~\cite{kwiatkowski2019NQ}, TriviaQA~\cite{joshi2017triviaqa}, and PopQA~\cite{mallen2022popqa}) and multi-hop reasoning (HotpotQA~\cite{yang2018hotpotqa}, 2WikiMultiHopQA~\cite{wikimultihop}, MuSiQue~\cite{trivedi2022musique}, and Bamboogle~\cite{press2022bambooogle}), reporting F1 and LLM-as-Judge (LasJ) metrics consistent with \citet{gao2025asearcher}. We provide implementation details in Appendix~\ref{app:details}.

\subsection{Main Results}

\paragraph{In-Domain Evaluation.}

We first evaluate CARL under the same configuration as training on in-domain knowledge question-answering datasets, using a local retrieval server. 
Table~\ref{tab:local-kb} reports both performance (F1, LasJ) and efficiency metrics ($|\mathcal{D}_\text{roll}|$, $|\mathcal{D}_\text{upd}|$), where the latter captures the number of actions performed during rollout and the number of actions used for policy updates, respectively. Overall, CARL consistently outperforms GRPO across all settings with a significantly reduced number of update actions.

For non-reasoning models, the 3B variant yields modest improvements due to the model’s limited capability -- trajectories tend to be short, and action-level rewards consequently provide less benefit.
On the 7B backbone, CARL's precise credit assignment enables longer agentic pipelines and yields over 2-point gains on both F1 and LasJ; although rollout cost increases, the substantial reduction in update actions keeps overall cost comparable, reflecting more rational resource allocation.

Considering the trade-off between performance and training cost, we conduct analytical experiments on the 4B reasoning model with a maximum of 10 actions. When the initial sample size is set to $N_0=1$, CARL-Lite achieves higher performance while using less than half the rollout cost and under 40\% of the update samples. Increasing the initial sample size to $N_0=8$ further enhances stability and diversity, enabling CARL to surpass GRPO by 1.4 points on average while still maintaining superior efficiency.
Extending the maximum actions to 32 further amplifies CARL's advantage, with performance gains increasing to 2.2 points over GRPO. 

These results demonstrate that CARL is a highly efficient RL algorithm for multi-turn search agents, achieving stronger performance through precise credit assignment while reducing training cost by eliminating redundant computation. Notably, the benefits amplify as models' foundational capability gets stronger and trajectories get longer, suggesting increasing practical value on more complex tasks.

\paragraph{Baseline Comparison.}

To validate the effectiveness of CARL, we reproduce three related methods that improve GRPO from the rollout strategy perspective: TreeRPO~\cite{yang2025treerpo}, TreeRL~\cite{hou2025treerl}, and ARPO~\cite{dong2025arpo}. Since these methods are originally designed for token-level tasks with different training configurations, we adapt them to multi-turn search tasks using the 4B reasoning backbone, ensuring an identical number of terminal states per group for fair comparison. Implementation details are provided in Appendix~\ref{app:related-implementation}. As shown in Table~\ref{tab:local-kb}, all three baselines achieve efficiency gains over GRPO through tree-structured partial sampling. However, through criticality-aware RL, CARL-Lite achieves comparable performance with lower computational cost. When computing resource consumption is comparable, CARL exhibits superior performance.

\paragraph{Out of Domain Evaluations.}
\begin{table}[t]
\centering

\caption{\textbf{Out-of-Distribution Evaluation.} We use a 4B reasoning model with a 32-action budget, reporting mean (Avg@4) and best-of-four (Pass@4) accuracy across 4 random seeds.}
\resizebox{\linewidth}{!}{
\begin{tabular}{l|cc|cc|cc}
\toprule
\multirow{2}{*}{Method} 
& \multicolumn{2}{c|}{GAIA} 
& \multicolumn{2}{c|}{Frames} 
& \multicolumn{2}{c}{xBench-DS}
\\
& Avg@4 & Pass@4 
& Avg@4 & Pass@4 
& Avg@4 & Pass@4 
\\
\midrule
Zero-Shot       & 27.2 & 44.7 & 44.5 & 63.6 & 36.2 & 58.0 \\
GRPO            & 30.6 & 49.5 & 53.2 & 71.2 & \textbf{46.0} & \textbf{71.0} \\
\rowcolor{blue!10}
CARL            & \textbf{32.5} & \textbf{54.4} & \textbf{57.1} & \textbf{75.8} & \textbf{46.0} & 70.0 \\ 
\bottomrule
\end{tabular}
}
\label{tab:search_results}
\end{table}

We evaluate the impact of CARL on the model's out-of-distribution capability. We replace the local database with online search and website access tools and test agents on three more challenging benchmarks. Following ASearcher~\cite{gao2025asearcher}, we evaluate each model with 4 random seeds and report both mean and best-of-four results.
As shown in Table~\ref{tab:search_results}, CARL achieves significantly higher performance on GAIA~\cite{mialon2023gaia} and Frames~\cite{krishna2025frames}, and comparable performance on xBench-DeepSearch~\cite{chen2025xbench}.
This demonstrates the advantage of CARL's criticality-aware learning strategy in enhancing decision-making and mitigating overfitting, leading to stronger generalization.

\subsection{Ablation Study}

Here we provide a breakdown analysis of CARL by ablating its key components: action-level advantage formulation, entropy-guided progressive rollout, and selective update on actions from high-criticality states. Results are shown in Table~\ref{tab:ablation}.

\paragraph{Action-Level Advantage Formulation.}
To isolate the effect of action-level advantage formulation, we design a variant (Exp. \#1) that replaces action-level rewards with outcome rewards. This variant collects trajectories in the same manner as CARL. Differently, each root-to-leaf chain is treated as an independent trajectory, and the outcome reward is uniformly assigned to all actions taken from critical states within that trajectory. For those actions that appear in multiple root-to-leaf chains, we include every corresponding instance in the update set, each paired with its respective outcome reward.
Compared with CARL (Exp. \#4), this variant yields weaker performance due to the noisy credit assignment, demonstrating the advantage of action-level advantage formulation.

\begin{table*}[t]
\centering
\small
\caption{\textbf{Ablation Results.} $|\mathcal{D}_\text{roll}|$ Avg. and $|\mathcal{D}_\text{upd}|$ Avg. denote the average number of actions performed during rollout phase and samples used for model update per task, respectively. Row \#4 (highlighted) represents the full CARL configuration. Best results are in bold.}
\label{tab:ablation}
\begin{tabular}{c|ccc|cc|ccc}
\toprule
\# & Rollout & Reward & Update & F1 Avg. & LasJ Avg. & $|\mathcal{D}_\text{roll}|$ Avg. & $|\mathcal{D}_\text{upd}|$ Avg. \\
\midrule
1 & Entropy-Guided & Outcome & High-Criticality & 52.9 & 55.0 &\textbf{65.8} & 45.5 \\
2 & Random & Action-Level & High-Criticality & 56.2 & 58.1 & 91.2 &\textbf{30.1} \\
3 & Entropy-Guided & Action-Level & All & 57.5 & 59.4 & 66.5 & 66.5 \\
\rowcolor{blue!10}
4 & Entropy-Guided & Action-Level & High-Criticality & \textbf{58.4} & \textbf{60.5} & 81.8 & 32.0 \\
\bottomrule
\end{tabular}
\end{table*}

\paragraph{Entropy-Guided Progressive Rollout.}
To verify the necessity of entropy-guided progressive rollout, we design a variant (Exp. \#2) by replacing CARL's forking algorithm with random selection. Compared with entropy-guided (Exp. \#4), random selection leads to a performance drop of 2.2 points in F1 and 2.4 points in LasJ. This demonstrates that entropy-guided progressive rollout plays a crucial role in CARL, as it directs exploration toward critical states rather than treating all states equally.

\paragraph{Criticality-Aware Selective Update.}
To assess the impact of excluding actions from low-criticality states during model update, we design a variant that retains all actions.
We supplement the advantage definition for actions from low-criticality states by letting them inherit the advantage of their preceding action.
In this way, they jointly contribute to the expected reward gain, analogous to the reward-sharing scheme used in GRPO.
Comparing the variant that keeps all actions (Exp. \#3) against CARL (Exp. \#4), we observe that CARL provides a clear performance gain of around 1 point for both metrics under the comparable computational cost. This indicates that CARL achieves better allocation of computational resources through focusing on actions from critical states.

\subsection{Analysis}
In this section, we provide further analysis to investigate two research questions: \textbf{RQ1:} How does CARL achieve significant improvement on OOD benchmarks? \textbf{RQ2:} What are the characteristics of high-criticality states?

\paragraph{Diversity Impact (RQ1).}
\begin{figure}
    \centering
    \includegraphics[width=1\linewidth]{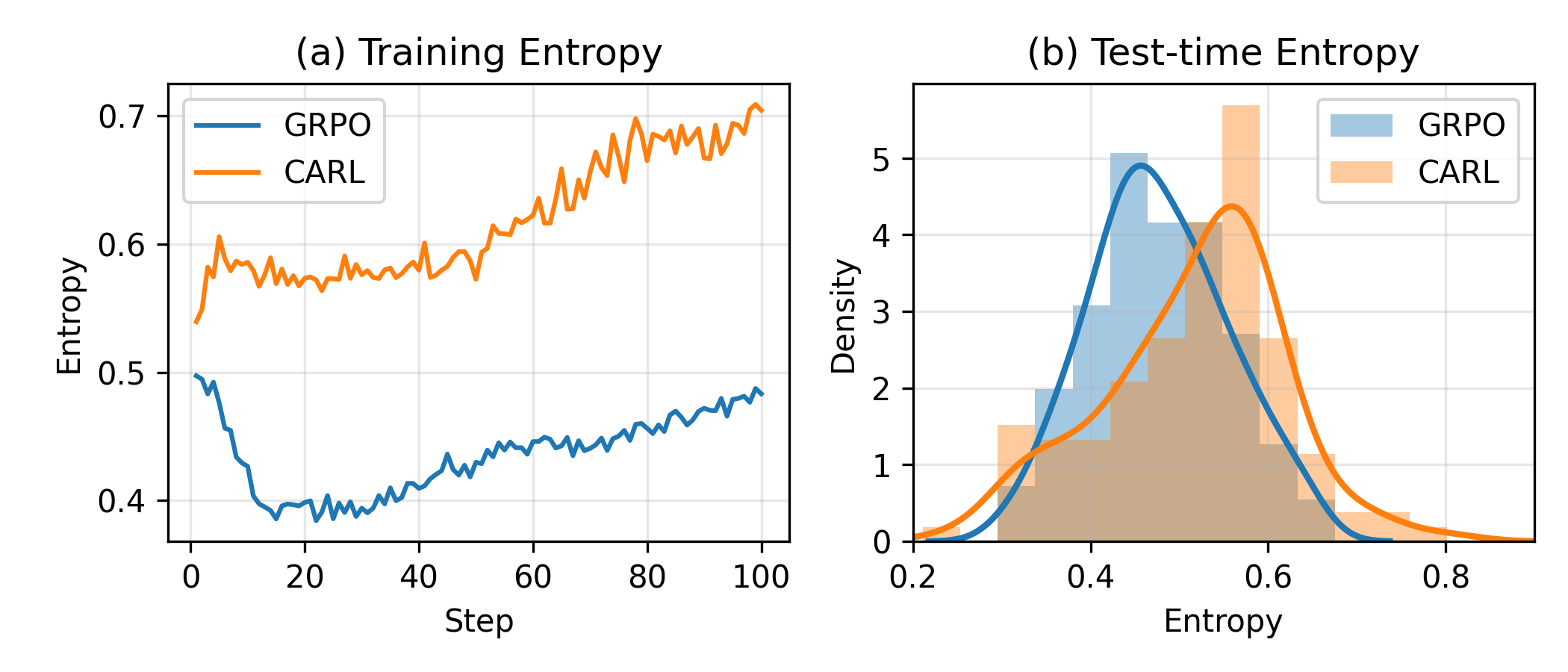}
    \caption{\textbf{Comparison of Entropy between CARL and GRPO.} CARL maintains consistently higher entropy than GRPO during training and evaluation, indicating stronger exploration capability.}
    \label{fig:entropy-test}
\end{figure}

Recent works~\cite{cheng2025reasoning, cui2025entropy, yue2025does} show that standard RL training will reduce model diversity and drive the policy to overly deterministic behavior. Diversity is important for agentic reasoning, as it enables the model to continuously reason and explore with the environment rather than jumping to conclusions prematurely.

Unlike GRPO, CARL selectively updates only actions taken from critical states instead of uniformly optimizing the entire trajectory. This conservative update scheme preserves model diversity.
As shown in Fig.~\ref{fig:entropy-test}(a), the policy entropy of CARL remains consistently higher than that of GRPO and continues to increase throughout training. In contrast, GRPO rapidly collapses to a lower-entropy regime and exhibits limited recovery, suggesting reduced exploration capacity as training proceeds. We also examine the entropy on unseen test datasets. As shown in Fig.~\ref{fig:entropy-test}(b), the entropy distribution of CARL is overall higher than that of GRPO, indicating that CARL maintains a higher-entropy policy and thereby has richer action diversity and stronger exploration potential. This explains CARL's advantage on OOD benchmarks.

\begin{figure}[t]
    \centering
    \includegraphics[width=1.0\linewidth]{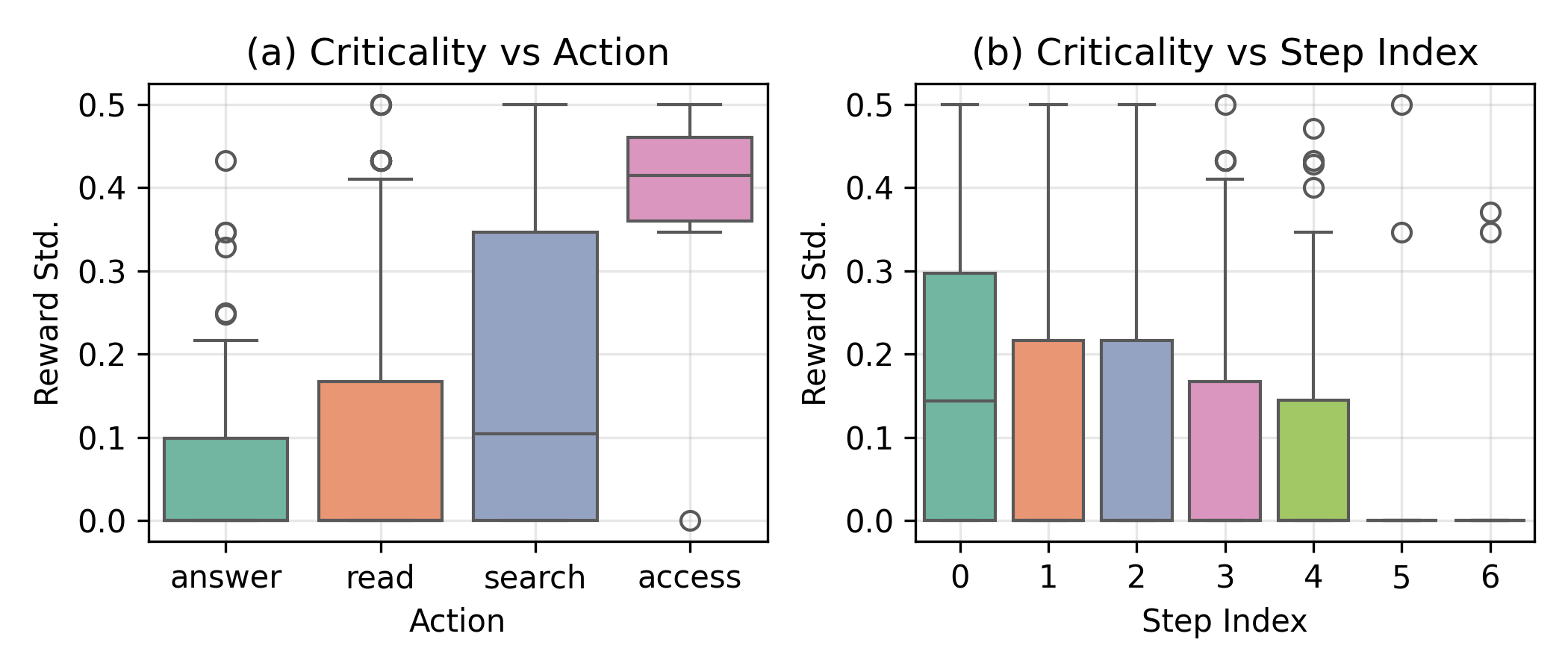}
    \caption{\textbf{Further Action Criticality Analysis.} Criticality distribution across (a) action types and (b) step positions. The distribution aligns with intuition but is not solely determined by type or position, necessitating per-action identification.}
    \label{fig:appendix-pipe}
\end{figure}
\paragraph{Further State Criticality Analysis (RQ2).}
We analyze state criticality across different action types and step positions. Fig.~\ref{fig:appendix-pipe}(a) shows that states preceding \textit{search} and \textit{access} actions exhibit higher criticality than those preceding \textit{read} and \textit{answer} actions. This aligns with the intuition that selecting search keywords and URLs is more critical than summarizing information or generating answers. Also, in Fig.~\ref{fig:appendix-pipe} (b), we can observe that the first few states the agent encounters generally have a greater impact on the outcome than subsequent ones. However, the outcome is neither equally determined by all states nor entirely determined by a single one, which verifies the necessity of dynamic critical state identification.

\section{Conclusion}
This paper analyzes the execution pipeline of multi-turn search agents and proposes CARL according to the discrepancy in state criticality. By focusing RL on actions taken from critical states, CARL effectively addresses two key limitations of GRPO: noisy credit assignment and redundant computation. As large reasoning models and long-horizon agentic reasoning tasks continue to gain prominence, the design philosophy of CARL, criticality-aware learning, offers a principled and efficient solution for RL in such settings. In future work, we plan to extend CARL to more challenging tasks involving ultra-long-horizon agentic reasoning and multi-agent systems, where the benefits of CARL are expected to be even more pronounced.

\bibliography{ref}
\bibliographystyle{icml2025}

\newpage
\appendix
\onecolumn

\section{Further Analysis}

\subsection{The Unbiasedness of Advantage Formulation}
\label{sec:unbiased}

CARL's advantage estimation is unbiased because it preserves the recursive structure of the Bellman equation. Specifically, the expected reward of a state $u$ is defined and estimated as:
\begin{equation}
\mathbb{E}[R(u)] = \sum_{v} p(v \mid u)\,\mathbb{E}[R(v)] = \frac{1}{N}\sum_{v_i \sim \pi_\theta(\cdot \mid u)} \mathbb{E}[R(v_i)]
\end{equation}

Since each child $v_i$ is sampled according to the policy $\pi_\theta$, the sample mean is an unbiased estimator of the true expectation. This property propagates recursively from leaf nodes up to the root, ensuring that the advantage $A(e) = \mathbb{E}[R(v)] - \mathbb{E}[R(u)]$ remains unbiased at every edge of the tree.

\paragraph{Bias Analysis of Alternatives.}

TreeRL~\cite{hou2025treerl} computes state values by averaging over all descendant leaf nodes (Eq.~\ref{eq:treerl}).
We now show that this value estimation is biased. Consider a state $s$ with two possible subsequent states $s_a$ and $s_b$. Suppose from $s_a$ we sample $n_a$ leaf nodes, each with reward $R_a$, and from $s_b$ we sample $n_b$ leaf nodes, each with reward $R_b$. The true state value is:
\begin{equation}
V^*(s) = \pi(a|s) \cdot R_a + \pi(b|s) \cdot R_b.
\end{equation}

TreeRL estimates the value by:
\begin{equation}
\hat{V}(s) = \frac{n_a \cdot R_a + n_b \cdot R_b}{n_a + n_b}.
\end{equation}

The difference between the expected value and true state value is
\begin{align}
\mathbb{E}[\hat{V}(s)] - V^*(s) = &\left(\mathbb{E}\left[\frac{n_a}{n_a + n_b}\right] - \pi(a|s)\right) R_a + \left(\mathbb{E}\left[\frac{n_b}{n_a + n_b}\right] - \pi(b|s)\right) R_b \\ = &\left(\mathbb{E}\left[\frac{n_a}{n_a+n_b}\right] - \pi(a|s)\right)(R_a - R_b).
\end{align}

Since $n_a$ and $n_b$ depend on the sampling process and tree expansion strategy, $\frac{n_b}{n_a + n_b} \ne \pi(a|s)$ in general, therefore $\mathbb{E}[\hat{V}(s)] - V^*(s) \ne 0$ cannot be guaranteed. Therefore, the advantage computed based on the biased value is also biased.

Another alternative method is provided by ARPO~\cite{dong2025arpo}, which computes trajectory-level advantages following GRPO, then averages them for shared tokens. For a token appearing in $d$ trajectories with final rewards $R_1, \ldots, R_d$, the shared advantage is:
\begin{equation}
\hat{A}^{\text{shared}} = \frac{1}{d}\sum_{i=1}^d \frac{R_i - \mu}{\sigma}= \frac{\bar{R} - \mu}{\sigma},
\end{equation}
where $\mu$ and $\sigma$ are the mean and standard deviation computed over the entire batch, and $\bar{R} = \frac{1}{d}\sum_i R_i$. The true advantage for taking action $a$ at state $s$ is defined as:
\begin{equation}
A^*(s,a) = \mathbb{E}[R \mid s, a] - \mathbb{E}[R \mid s].
\end{equation}

ARPO's estimator suffers from two sources of bias. First, the baseline $\mu$ is computed on batch level. We have $\mathbb{E}[\mu] \neq V^*(s)$, unless all trajectories in the batch originate from the same state $s$, which is generally not the case. 

Second, the sample mean $\bar{R}$ is not an unbiased estimator of $\mathbb{E}[R|s,a]$ when the $d$ trajectories passing through the shared token have imbalanced continuations, just as discussed before. Suppose after taking action $a$ at state $s$, the resulting state $s'$ branches into substates $s'_1$ and $s'_2$ with $\pi(\cdot|s') = 0.5$ for each. If $n_1$ trajectories continue through $s'_1$ (with reward $R_1$) and $n_2$ through $s'_2$ (with reward $R_2$), then:
\begin{equation}
\bar{R} = \frac{n_1 R_1 + n_2 R_2}{n_1 + n_2}
\end{equation}
\begin{equation}
\mathbb{E}[\bar{R}] = \mathbb{E}\left[\frac{n_1}{n_1+n_2}\right] R_1 + \mathbb{E}\left[\frac{n_2}{n_1+n_2}\right] R_2 \neq 0.5 R_1 + 0.5 R_2 = \mathbb{E}[R|s,a]
\end{equation}

The inequality holds whenever $n_1$ and $n_2$ are not identically distributed, which is clearly not guaranteed. Since the two biases arise from independent sources---batch composition and subtree sampling structure---they obviously cannot cancel, resulting in biased advantage estimates.

The unbiasedness of CARL's estimator provides important practical benefits: it allows flexible adjustment of the sampling tree structure without introducing systematic errors in advantage estimation, which makes the uncertainty-guided progressive rollout algorithm possible. This decouples the exploration strategy from estimation correctness, enabling more efficient use of computational resources.

\subsection{Inference Efficiency}

We measure the inference efficiency and summarize the results in Table~\ref{tab:inference}. Across all settings, CARL-trained models achieve comparable or lower per-action token usage compared to GRPO, indicating that CARL does not introduce unnecessary verbosity.
For larger backbones such as the 7B non-reasoning variant, CARL further demonstrates its ability to support extended multi-step execution, enabling significantly longer action sequences and yielding performance gains exceeding 2 points.
These advantages stem from action-level reward modeling and critical-state-focused model update.

\begin{table}[t]
\centering
\caption{\textbf{Inference Efficiency Comparison.} Token denotes the average number of tokens generated per action; Action denotes the average number of actions executed by the model.}
\label{tab:inference}
\begin{tabular}{l|cc|cc|cc}
\toprule
& \multicolumn{2}{c|}{\textbf{3B Non-Reasoning}} & \multicolumn{2}{c|}{\textbf{7B Non-Reasoning}} & \multicolumn{2}{c}{\textbf{4B Reasoning}} \\
& GRPO & CARL & GRPO & CARL & GRPO & CARL \\
\midrule
Token & 50.85 & 43.67 & 40.67 & 40.90 & 962.07 & 807.11 \\
Action & 3.40 & 3.27 & 4.30 & 7.52 & 4.90 & 4.56 \\
\midrule
F1 & 33.3 & 34.0 & 46.3 & 48.5 & 57.0 & 58.4 \\
LasJ  & 32.5 & 33.2 & 48.3 & 50.5 & 59.5 & 60.5 \\
\bottomrule
\end{tabular}
\end{table}

\subsection{Computational Resource}

All of our experiments are conducted on 8$\times$A100 GPUs. We provide a comparison of computational resource usage in Table~\ref{tab:gpu-hours}. CARL-Lite achieves the lowest GPU hours (253.17) by significantly reducing both rollout and update costs, while still outperforming GRPO on both metrics. CARL, with a larger initial sample size, increases rollout cost to a level comparable to GRPO, delivering larger performance improvements while still requiring fewer computational resources. These results confirm CARL's efficiency advantages.

\begin{table}[t]
\centering
\caption{\textbf{Computational Resource Comparison.} We report GPU hours for training using the Qwen3-4B reasoning model as backbone, with all experiments conducted on A100.}
\label{tab:gpu-hours}
\begin{tabular}{l|cc|c|cc}
\toprule
Method & $|\mathcal{D}_\text{roll}|$ & $|\mathcal{D}_\text{upd}|$ & GPU Hours & F1 Avg. & LasJ Avg.\\
\midrule
GRPO & 80.9 & 80.9 & 423.05 & 57.0 & 59.5 \\
CARL-Lite & 39.8 & 30.5 & 253.17 & 57.6 & 59.6 \\ 
CARL & 81.8 & 32.0 & 403.92 & 58.4 & 60.5\\
\bottomrule
\end{tabular}
\end{table}

\subsection{Case Study}
\begin{table}

\centering
\caption{Case Study: Pipeline Comparison Between GRPO-Trained and CARL-Trained Agents. Ground Truth: John Leventhal.}
    \label{tab:case}
\begin{tabular}{c|p{7.8cm}|p{7.8cm}}

\toprule

\# & \textbf{GRPO} & \textbf{CARL} \\

\midrule
\multicolumn{3}{c}{\textbf{Question:} Who is the spouse of the performer of song Hold On (Rosanne Cash Song)?} \\
\midrule

1 &
\textbf{Think:} The question asks for the spouse of the performer of the song "Hold On" specifically labeled as a ``Rosanne Cash Song.'' ... I need to confirm whether Rosanne Cash is the performer of this song ... Without knowing the definitive performer, I cannot proceed to find their spouse ... the next logical step is to search for the song's performer ...

\textbf{Search:} Who performed the song Hold On (Rosanne Cash Song)?
&
\textbf{Think:} I need to find out who the spouse of the performer of the song ``Hold On'' (described as a Rosanne Cash song) is. ... I'll start by searching to verify the performer of ``Hold On'' associated with Rosanne Cash to make sure I have the right person before looking for their spouse.

\textbf{Search:} Who performs the song Hold On (Rosanne Cash song)?
\\
\midrule

2 &
\textbf{Think:} After reviewing the search results, I confirmed that Rosanne Cash is indeed the performer of ``Hold On'' ... Doc 1: ``Hold On'' is a song written and recorded by American country music artist Rosanne Cash ... The critical information about her spouse comes from Doc 5, which states: ``Crowell and Cash married in 1979 ...'' ... this almost certainly refers to Rodney Crowell ... Rosanne Cash's spouse is Rodney Crowell ...
&
\textbf{Think:} After reviewing the search results, I've identified that Rosanne Cash is the performer of the song ``Hold On'' (as indicated in Doc 1, ... I need to find information about Rosanne Cash's spouse. My next step should be to search for ``Rosanne Cash spouse'' to identify who her spouse is.
\\
\midrule

3 &
\textbf{Think:} Rosanne Cash is confirmed as the performer of the song ``Hold On'' ... ``Crowell and Cash married in 1979'' ... it's clear that ``Crowell'' refers to Rodney Crowell ... There's no indication in the search results of any other potential spouse ... the answer is sufficiently supported ...

\colorbox{blue!10}{\textbf{Answer:} Rodney Crowell}
&
\textbf{Think:} Okay, now that I've confirmed from the search results that Rosanne Cash is the performer of the song ``Hold On'', my next step is to find information about her spouse. ... I'll generate a search query for this information.

\textbf{Search:} Rosanne Cash spouse

\\
\midrule

4 & (Finished) &

\textbf{Think:} From Document 1 (Wikipedia page for Rosanne Cash), it clearly states: ``She married her second husband, John Leventhal, in 1995, and they have one son, Jakob ... Document 5 also provides confirmation: ``Rosanne was married to country superstar, Rodney Crowell, from 1979 to 1992. She then married musician, John Leventhal.'' ... the answer is John Leventhal.

\\
\midrule

5 & (Finished) &

\textbf{Think:} After analyzing the search results for ``Rosanne Cash spouse'', I found clear information from multiple sources confirming that Rosanne Cash married John Leventhal in 1995. 

\colorbox{blue!10}{\textbf{Answer:} John Leventhal}
\\
\bottomrule

\end{tabular}
\end{table}

We present an example of how CARL succeeds in accomplishing difficult tasks in Table~\ref{tab:case}.  In this case, the GRPO-trained agent commits to an answer immediately after encountering a limited piece of evidence -- ``Crowell and Cash married in 1979''. In contrast, the CARL-trained agent exhibits a more cautious and plan-driven strategy, performing additional searches to explicitly verify the spouse information from multiple sources. Through this process, it found more comprehensive evidence -- ``Cash married her second husband, John Leventhal, in 1995'' -- and answers correctly.
In this case, the state preceding the third action is the most critical, as the decision of the model at this point determines whether it can obtain sufficient information for a correct answer.

\subsection{Statistical Significance Analysis}
\label{app:significance}

To verify that the improvements reported in Table~\ref{tab:local-kb} are not attributable to random variation across runs, We conduct paired t-tests on the two configurations of the 4B Reasoning model: with a maximum of 10 actions and a maximum of 32 actions. For each configuration, we run 4 independent evaluations with different random seeds for both the GRPO baseline and our method, and apply a paired t-test on the per-seed F1 and LasJ scores.

Table~\ref{tab:ttest} summarizes the t-statistics and corresponding p-values. Across both configurations and both metrics, all p-values fall well below the conventional significance threshold of $0.05$. These results provide evidence that the gains reported in Table~\ref{tab:local-kb} reflect genuine methodological improvements over the GRPO baseline.

\begin{table}[h]
\centering
\caption{Paired t-test results comparing our method against the baseline on the 4B Reasoning model under two action-budget settings.}
\label{tab:ttest}
\begin{tabular}{lcc}
\toprule
\textbf{Metric} & \textbf{4B-max10} & \textbf{4B-max32} \\
\midrule
F1 --- $t$         & 5.89   & 8.878  \\
F1 --- $p$         & 0.0098 & 0.0125 \\
\midrule
LasJ --- $t$       & 6.32   & 13.234 \\
LasJ --- $p$       & 0.0080 & 0.00566 \\
\bottomrule
\end{tabular}
\end{table}

\section{Discussion of Limitations}\label{app:limitation}

\paragraph{Baseline Capability Requirement.}
CARL is built on the assumption that model uncertainty correlates with state criticality. If the model confidently makes an incorrect decision at a state, CARL will be less effective than GRPO, as it will not explore alternative choices at that state. In other words, CARL yields a higher performance ceiling only when the model possesses sufficient baseline capability. Under this condition, uncertainty serves as a reliable signal for identifying critical states, allowing CARL to selectively refine the actions taken there and achieve stronger overall performance.

This is empirically supported by our results in Table~\ref{tab:local-kb}: CARL demonstrates larger improvements over GRPO on stronger base models (e.g., 7B vs. 3B non-reasoning models), where the model has greater foundational capacity to express more meaningful possibilities at key decision points.

Therefore, for more challenging tasks where agents exhibit low zero-shot performance, a cold-start phase via supervised fine-tuning (SFT) is necessary to establish basic competence before CARL can be applied for further unsupervised improvement.

\paragraph{Experiments.}
Due to limited computational resources, we have not yet extended CARL to larger reasoning models such as QwQ-32B, or evaluated it on more complex agentic benchmarks~\cite{wei2025browsecomp, workarena2024} and multi-agent frameworks~\cite{ye2025maslab}. We will validate our approach on them in future work.

\section{Implementation Details}\label{app:details}

\subsection{Uncertainty-Guided Progressive Rollout Algorithm}\label{app:rollout}

Here we provide the implementation details of the uncertainty-guided progressive rollout algorithm. As described in Algorithm~\ref{alg:rollout}, the rollout process consists of two phases. In the first phase, we generate $N_0$ trajectories from scratch to establish a basic set of candidate nodes. Notably, actions in this phase are not included in training. In the second phase, we iteratively select the state with the lowest action density to fork from.

\begin{algorithm}[h]
\caption{Uncertainty-Guided Progressive Forking}
\label{alg:rollout}
\begin{algorithmic}[1]
\STATE {\bfseries Input:} Policy $\pi_\theta$, root state $s_0$, total rollouts $N$, initial sample size $N_0$
\STATE {\bfseries Output:} Sampled tree $\mathcal{T}$

\STATE Initialize $\mathcal{T}\leftarrow\{s_0\}$, state buffer $\mathcal{S}\leftarrow\emptyset$

\FOR{$i = 1$ to $N_0$}
    \STATE $\tau \leftarrow \textsc{RolloutFrom}(s_0;\pi_\theta)$
    \FOR{each $(s,a)$ in $\tau$}
        \STATE $h \leftarrow \textsc{Entropy}(a)$
        \STATE Add $(s,h,1)$ to $\mathcal{S}$ 
        \RComment{record (state, uncertainty, count)}
    \ENDFOR
\ENDFOR

\FOR{$i = 1$ to $N$}
    \STATE Select $(\hat{h}, \hat{s}, \hat{n})$ with biggest $\hat{h}/\hat{n}$ from $\mathcal{S}$ 
    
    \RComment{pick the state with the lowest action density}
    \STATE $\tau \leftarrow \textsc{RolloutFrom}(\hat{s};\pi_\theta)$
    \STATE $T \leftarrow T + 1$
    \FOR{each $(s',a')$ in $\tau$}
        \STATE $h' \leftarrow \textsc{Entropy}(a')$
        \IF{$s' = \hat{s}$} 
        
            \STATE $\hat{h} \leftarrow \dfrac{h' + \hat{h} \cdot \hat{n}}{\hat{n} + 1}, \quad \hat{n} \leftarrow \hat{n} + 1$
            \STATE Update $(\hat{h}, \hat{s}, \hat{n})$ in $\mathcal{S}$ 
                \RComment{Update uncertainty}
        \ELSE
            \STATE Add $(s', h', 1)$ to $\mathcal{S}$ 
            \RComment{insert new candidate}
        \ENDIF
    \ENDFOR
\ENDFOR

\end{algorithmic}
\end{algorithm}

\subsection{Tree-Based Advantage Estimation Algorithm}

Here we provide the implementation details of the tree-based advantage estimation algorithm. As described in Algorithm~\ref{alg:tree_advantage}, the calculation process is implemented by two deep first searches.
In the first pass (Bottom-up Value Estimation), we propagate rewards from leaf nodes to the root. In the second pass (Top-down Advantage Collection), we compute the advantage $\mathcal{A}(u)$ for each action. This process yields a set of advantage-action-paired samples $\mathcal{D}$ for policy update.

\begin{algorithm}[h]
\caption{Tree-based Advantage Estimation}
\label{alg:tree_advantage}
\begin{algorithmic}[1]
\STATE {\bfseries Input:} Reasoning Tree $\mathcal{T}$ with root node $u_{\text{root}}$
\STATE {\bfseries Output:} Advantage training dataset $\mathcal{D}$

\STATE Initialize $\mathcal{D} \leftarrow \emptyset$

\RComment{Phase 1: Bottom-up Value Estimation (Post-Order DFS)}
\STATE \textbf{Procedure} \textsc{EstimateValue}($u$) 
    \IF{$u$ is a leaf node}
        \STATE $\mathcal{V}(u) \leftarrow R(u)$  \RComment{Final reward from environment}
    \ELSE
        \FOR{each child $v$ in $u.\text{children}$}
            \STATE \textsc{EstimateValue}($v$)
        \ENDFOR
        \STATE $\mathcal{V}(u) \leftarrow \frac{1}{|u.\text{children}|} \sum_{v} \mathcal{V}(v)$ \quad \RComment{Mean value aggregation}
    \ENDIF
\STATE \textbf{End Procedure}

\RComment{Phase 2: Top-down Advantage Collection (Pre-Order DFS)}
\STATE \textbf{Procedure} \textsc{CollectAdvantage}($u, \mathcal{V}_{\text{parent}}$)
    \IF{$u \neq u_{\text{root}}$}
        \STATE $\mathcal{A}(u) \leftarrow \mathcal{V}(u) - \mathcal{V}_{\text{parent}}$
        \STATE Add $(\text{seq}_u, \mathcal{A}(u))$ to $\mathcal{D}$ \quad \RComment{Record sequence and local advantage}
    \ENDIF
    \FOR{each child $v$ in $u.\text{children}$}
        \STATE \textsc{CollectAdvantage}($v, \mathcal{V}(u)$)
    \ENDFOR
\STATE \textbf{End Procedure}

\RComment{Main Execution}
\STATE \textsc{EstimateValue}($u_{\text{root}}$)
\STATE \textsc{CollectAdvantage}($u_{\text{root}}, 0$)

\end{algorithmic}
\end{algorithm}

\subsection{Training Details}\label{app:train-details}
We adopt Qwen2.5-3B and Qwen2.5-7B~\cite{qwen2.5} as backbones for non-reasoning models, and Qwen3-4B~\cite{qwen3} as the backbone for reasoning models. We use a valid subset of ASearcher-base with invalid questions excluded for training stability. The reward function combines a format reward with an F1-based answer reward. Due to computational constraints, we adopt different hyperparameter configurations for the two model types: non-reasoning models are trained with a batch size of 128 and a learning rate of $5\times10^{-6}$; reasoning models use a batch size of 64 and a learning rate of $1\times10^{-5}$. All models are trained for 100 optimization steps, except for 4B reasoning models with a maximum of 32 actions, which are trained for 50 steps due to computational resource constraints.

\subsection{Evaluation Metrics}\label{app:eval-details}
Following ASearcher~\cite{gao2025asearcher}, we adopt two complementary metrics to evaluate answer quality from different perspectives: F1 as a rule-based metric, and LLM-as-Judge (LasJ) as a model-based metric. F1 measures the word-level overlap between predictions and reference answers, computed as the harmonic mean of precision and recall. LasJ evaluates correctness by prompting GPT-5-nano-2025-08-07 to compare model predictions against reference answers, outputting a binary judgment.
The evaluation prompt is shown in Fig.~\ref{fig:judge-prompt}.

\begin{figure}[t]
\centering
\begin{tcolorbox}[
    colback=gray!3,
    colframe=gray!50,
    boxrule=0.5pt,
    arc=2pt,
    left=6pt,
    right=6pt,
    top=6pt,
    bottom=6pt,
    width=\columnwidth
]
\small\ttfamily
You are an evaluation assistant. Please determine if the predicted answer is equivalent to the labeled answer.\\[0.8em]
Question: \textit{\{question\}}\\[0.4em]
Labeled Answer: \textit{\{gt\_answer\}}\\[0.4em]
Predicted Answer: \textit{\{pred\_answer\}}\\[0.8em]
Did the model give an answer **equivalent** to the labeled answer? Please respond with "Correct" if they are equivalent, or "Incorrect" if they are not equivalent.\\[0.8em]
The output should in the following json format:\\
\begin{lstlisting}
```json
{
    "rationale": <your rationale>,
    "judgement": "Correct" or "Incorrect"
}
```
\end{lstlisting}
\end{tcolorbox}
\caption{Prompt template for LLM-based answer evaluation. Placeholders in \textit{italics} are replaced with actual values during evaluation.}
\label{fig:judge-prompt}
\end{figure}
\subsection{Related Baselines}\label{app:related-implementation}

\paragraph{TreeRPO.} TreeRPO~\cite{yang2025treerpo} focuses on math reasoning tasks and leverages tree-structured sampling to provide rewards for intermediate reasoning steps. It splits the reasoning process into fixed-length segments, converting single-step reasoning into multi-step reasoning. During the rollout phase, it forks at the end of each segment, resulting in a full tree structure. Samples sharing the same parent node are then grouped together for GRPO's group-relative reward computation.
In its original design, TreeRPO samples a full 8-ary tree with a maximum depth of 3. However, the sample size grows exponentially with tree depth, resulting in unaffordable computational costs for agentic tasks. To achieve a fair comparison, we perform partial expansion by randomly selecting a limited number of nodes to fork, ensuring that the number of leaf nodes remains consistent across all methods.

\paragraph{TreeRL.} TreeRL~\cite{hou2025treerl} also focuses on math reasoning tasks and further removes the group constraint. For intermediate reward computation, it proposes a hybrid advantage calculation algorithm that combines local advantages and global advantages, which can be expressed as:
\begin{equation}\label{eq:treerl}
V(s_n) \;=\; \frac{1}{\lvert L(s_n) \rvert}
\sum_{l \in L(s_n)} R(l)
\end{equation}
\begin{equation}
R(s_n) \;=\; \underbrace{\lvert L(s_n) \rvert^{-1/2}}_{\text{Re-weight Factor}}
\cdot
\Big(
\underbrace{V(s_n) - V(\text{root})}_{\text{Global Advantage}}
\;+\;
\underbrace{V(s_n) - V(p(s_n))}_{\text{Local Advantage}}
\Big)
\end{equation}
where $s_n$ denotes a reasoning step, $L(s_n)$ denotes the set of leaf nodes descending from $s_n$, and $p(s_n)$ denotes the parent of node $s_n$. 

\paragraph{ARPO.} ARPO~\cite{dong2025arpo} applies tree-structured sampling to multi-turn agentic reasoning by branching reasoning paths if entropy increases after tool calling. Specifically, it computes a normalized entropy change $\Delta H_t = \text{Normalize}(H_t - H_{\text{initial}})$, where $H_t$ is the step-level entropy after tool-call step $t$. The partial sampling probability is then determined by:
\begin{equation}
P_t = \alpha + \beta \cdot \Delta H_t, \quad \text{Action}(P_t) = 
\begin{cases}
\text{Branch}(Z), & \text{if } P_t > \tau \\
\text{Continue}, & \text{otherwise}
\end{cases}
\end{equation}
where $\alpha$ is a base sampling probability, $\beta$ is an entropy weight, and $\tau$ is a predefined threshold. When $P_t$ exceeds $\tau$, ARPO branches $Z$ additional partial reasoning paths from the current node. For advantage computation, ARPO first calculates the group-relative advantage $\hat{A}_{i,t}$ for each leaf node following GRPO. For shared tokens that appear in $d$ trajectories, it uses an averaged advantage: $\hat{A}_{i,t}^{\text{shared}} = \frac{1}{d}\sum_{i=1}^{d}\hat{A}_{i,t}$.

\end{document}